\documentclass[sigconf,screen]{acmart}

\usepackage{multirow}
\usepackage[table]{xcolor}
\usepackage{algorithm}
\usepackage{algpseudocode}
\usepackage{stfloats}

\AtBeginDocument{%
  }

\copyrightyear{2026}
\acmYear{2026}
\setcopyright{cc}
\setcctype{by-nc-nd}
\acmConference[ICCAD '26]{IEEE/ACM International Conference on Computer-Aided Design}{November 08--12, 2026}{San Jose, CA, USA}
\acmBooktitle{IEEE/ACM International Conference on Computer-Aided Design (ICCAD '26), November 08--12, 2026, San Jose, CA, USA}
\acmDOI{10.1145/3831252.3834176}
\acmISBN{979-8-4007-2873-0/2026/11}

\begin{document}

\title{Flash-CNNCap: Capacitance Extraction via Image Mapping}

\author{Hector R. Rodriguez, ~ Jiechen Huang, ~ Wenjian Yu$^*$}
\thanks{This work was supported by National Science and Technology Major Project (2021ZD0114703). $^*$Corresponding author.}
\affiliation{
Dept. Computer Science \& Tech., BNRist, State Key Laboratory of Cryptography and Digital Economy Security, Tsinghua Univ., Beijing, China
\country{}
}




\begin{abstract}
We present Flash-CNNCap, a CNN-based capacitance extractor that reformulates full-matrix capacitance prediction as image-to-image regression over spatial contribution maps.  Prior scalar CNN-based extractors require $O(n^{2})$ forward passes to recover all pairwise capacitances in a window with $n$ conductors. Flash-CNNCap replaces the scalar target with dense contribution maps: a total-capacitance model and a master-conditioned coupling model each predict a spatial map that is reduced to conductor-level values through mask aggregation, cutting full-matrix reconstruction to $O(n)$ passes.  The resulting totals and symmetrized pairwise couplings define the corresponding Maxwell-style capacitance matrix under the standard off-diagonal sign convention.  The maps are learned from conductor-level labels without per-pixel supervision.  An ablation study over 13 model configurations 
selects a U-Net that matches ResNet baselines on total capacitance (1.5--3.1\% MARE) and achieves the strongest coupling accuracy (3.0--4.6\% MARE) across all evaluated CapBench subsets, with a $17.5\times$ full-matrix speedup on windows with 134 conductors on average.  A deployed pipeline reads Design Exchange Format (DEF) geometry and writes Standard Parasitic Exchange Format (SPEF) output, processing 1{,}024 windows in 51.23\,s with a $4.4\times$ speedup over OpenRCX on the same benchmark.  Code and trained models are available at \url{https://github.com/THU-numbda/flash-cnncap}.
\end{abstract}

\ccsdesc[500]{Hardware~Metallic interconnect}
\ccsdesc[500]{Hardware~3D integrated circuits}
\ccsdesc[500]{Computing methodologies~Neural networks}

\keywords{capacitance extraction, parasitic extraction, electronic design automation, convolutional neural networks}

\begin{teaserfigure}
\centering
  \includegraphics[width=\textwidth]{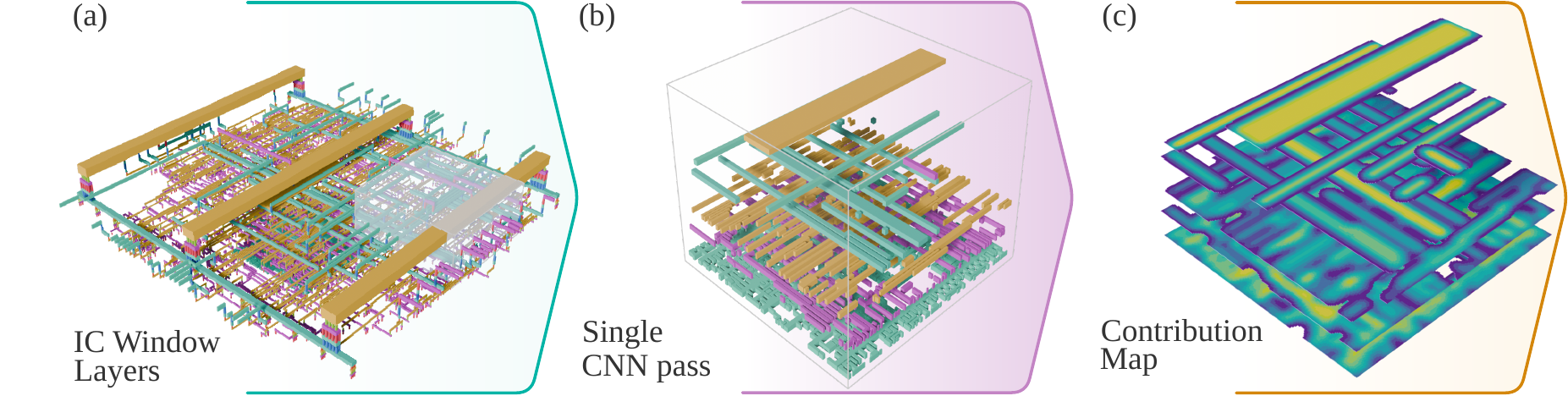}
  \caption{Flash-CNNCap overview: (a) IC structure, (b) single CNN pass on extraction window, and (c) contribution map.}
  \Description{Three-panel overview showing (a) IC structure in an extraction window, (b) a single CNN pass, and (c) the contribution map.}
  \label{fig:teaser}
\end{teaserfigure}

\maketitle

\section{Introduction}

Parasitic capacitance between interconnect wires is a key determinant of timing, signal integrity, and power in modern integrated circuits~\cite{Bakoglu1985,Bohr1995,yu2005capacitance,Yu2021,yu2012efficient,zhang2008efficient}.  Accurate extraction of the full capacitance matrix across a routed design is therefore a prerequisite for reliable sign-off.  Numerical field solvers such as FastCap~\cite{nabors1991fastcap}, QBEM~\cite{yu2004preconditioned} and RWCap~\cite{yu2013rwcap} deliver high accuracy but are computationally expensive, while rule-based extractors built on analytical models and pattern libraries~\cite{choudhury1995analytical,cong1997practical} are fast but sacrifice accuracy.  OpenRCX~\cite{OpenRCX2021} is a practical open-source implementation of this general rule-based approach.  Machine-learning extractors can offer near-solver accuracy at near-rule-based speed.  However, existing methods have not fully addressed the quadratic cost of reconstructing the full capacitance matrix.

CNN-Cap~\cite{yang2022cnncap} showed that ResNet models operating on layer-wise rasterized layouts can predict capacitance accurately.  However, its image-to-scalar formulation requires $n + n(n-1)/2$ forward passes to reconstruct the full capacitance matrix of a window with $n$ conductors, incurring $O(n^{2})$ cost that dominates runtime for large windows.  Other recent approaches explore different representation and preprocessing trade-offs, including graph-based GNN-Cap~\cite{liu2024gnncap} and point-cloud-based PCT-Cap~\cite{cai2024pctcap}, but none of these methods removes the quadratic reconstruction cost while preserving the raster formulation and demonstrating end-to-end deployment.

This paper presents Flash-CNNCap, which preserves the raster-based representation used in CNN-Cap but replaces scalar regression with image-to-image prediction of spatial contribution maps.  A total-capacitance model predicts one contribution map per window, and a master-conditioned coupling model predicts contributions for all other conductors visible in the window in a single pass, reducing full-matrix reconstruction to $O(n)$ evaluations.  These outputs define the corresponding Maxwell-style capacitance matrix by placing total capacitances on the diagonal and the negatives of the symmetrized pairwise couplings on the off-diagonal entries.  We evaluate Flash-CNNCap on the CapBench NanGate45 and Sky130HD benchmark subsets~\cite{rodriguez2026capbench}, which are built on FreePDK45~\cite{freepdk45} and the SKY130 open PDK~\cite{skywater130}, respectively, and compare against ResNet-34 and ResNet-50 baselines~\cite{He2016ResNet} retrained with an updated optimization recipe.

Our contributions are:
\begin{enumerate}
\item An image-to-image reformulation that reduces full-matrix CNN inference from $O(n^{2})$ to $O(n)$ passes through dense contribution maps and conductor-mask aggregation.
\item A controlled architecture study across 13 dense-prediction backbones, selecting a U-Net~\cite{ronneberger2015unet} that achieves the best coupling-accuracy--throughput trade-off.
\item Demonstration that the selected model matches ResNet baselines on total capacitance, achieves the strongest coupling accuracy on all evaluated CapBench subsets, and delivers a $17.5\times$ full-matrix speedup on large windows.
\item A GPU-accelerated DEF-to-SPEF extraction pipeline that processes 1{,}024 large windows in 51.23\,s, $4.4\times$ faster than OpenRCX on the same benchmark.
\end{enumerate}

\section{Related Work}

Capacitance extraction methods span a wide accuracy--throughput range.  At one extreme, numerical field solvers such as FastCap~\cite{nabors1991fastcap} and floating random walk (FRW) solvers such as RWCap~\cite{yu2013rwcap} achieve high accuracy but at substantial compute cost.  Later FRW refinements improve transition efficiency~\cite{yang2020floating,huang2024enhancing,rodriguez2026deeprwcap} without fundamentally reducing that cost.  At the other extreme, rule-based extractors approximate capacitances with analytical cross-section models and precharacterized pattern libraries~\cite{choudhury1995analytical,cong1997practical}.  OpenRCX~\cite{OpenRCX2021} is a practical open-source implementation of this regime.  These methods are fast and widely deployable, but their accuracy depends on the fidelity of the model library and can degrade on geometries or process conditions not covered by it, particularly for coupling capacitance.  Several machine-learning approaches have been proposed to bridge the accuracy--throughput gap between field solvers and rule-based extractors.

\paragraph{CNN-Cap~\cite{yang2022cnncap}} This CNN-based extractor rasterizes routed geometry into per-layer \emph{density maps}, where each pixel stores a continuous value representing the fractional area occupied by conductors, and trains a ResNet backbone to predict conductor-level capacitances with high accuracy.  However, its image-to-scalar formulation requires $n + n(n-1)/2$ forward passes to reconstruct the full capacitance matrix for a window with $n$ conductors, resulting in an $O(n^{2})$ cost that dominates runtime as conductor counts grow.

\paragraph{GNN-Cap~\cite{liu2024gnncap}} This method represents interconnect geometry as a graph of cuboid nodes and uses a graph convolutional network to predict total and coupling capacitance in a single forward pass, avoiding the quadratic pass count.  However, the graph construction requires cuboid decomposition, distance-threshold edge construction, and virtual-edge heuristics, adding geometry-preprocessing overhead that grows with layout complexity.  GNN-Cap also requires per-block capacitance labels from a field solver for each training sample, increasing dataset-construction effort relative to the conductor-level labels used by raster-based methods.

\paragraph{PCT-Cap~\cite{cai2024pctcap}} This method formulates extraction as a point-cloud task and applies a point cloud transformer, also reporting high accuracy.  Because the number of sampled points grows rapidly with window size, the quadratic attention cost of the transformer becomes a bottleneck for large windows.  PCT-Cap also retains the scalar prediction target, inheriting the same $O(n^{2})$ reconstruction cost as CNN-Cap for full-matrix extraction.

\paragraph{Dense prediction in EDA} Dense image-to-image prediction has appeared in related EDA tasks but not yet for capacitance extraction.  A WGAN-GP~\cite{lamichhane2021generative,gulrajani2017wgan} regresses finite-element electric-potential maps for back-end-of-line (BEOL) interconnect tiles, from which electric field and time-dependent dielectric breakdown (TDDB) lifetime are derived.  The paper reports over 200$\times$ speedup relative to the COMSOL solver on a single synthesized 32\,nm layout.  However, the training labels still require full COMSOL solves to generate pixel-wise potential maps, and the evaluation is limited to a single layout and one technology.  More fundamentally, the 2D per-layer formulation analyzes each metal layer in isolation, so inter-layer coupling capacitance is not captured.  Although per-pixel potential maps could in principle be post-processed to obtain capacitance, the missing vertical dimension makes such an extension non-trivial.

\paragraph{Other approaches} NAS-Cap~\cite{li2024nas} extends CNN-Cap with neural architecture search but retains its scalar prediction target.  Hybrid flows~\cite{abouelyazid2022hybrid} select among rule-based, neural-network, and field-solver engines, but remain bounded by the component extractors' limitations.  Pre-characterization-based methods~\cite{li2020precharacterization} use machine learning within a pattern-driven workflow, tying accuracy to the coverage of the pattern library.  RMLP-Cap~\cite{zhou2026rmlpcap} automates acquisition of parasitic-extraction (PEX) windows, feature extraction, and dataset generation with learned per-pattern models, but the per-pattern formulation ties generalization more closely to the characterized pattern space.

\section{Methodology}

\begin{figure}[h]
  \centering
  \includegraphics[width=\linewidth]{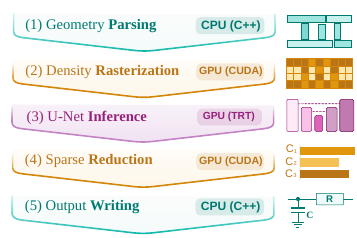}
  \caption{Flash-CNNCap pipeline from routed DEF input to SPEF output.}
  \Description{Block diagram of the deployed runtime stack from DEF parsing through GPU-resident rasterization, TensorRT inference, sparse reduction, and SPEF emission.}
  \label{fig:end_to_end}
\end{figure}

This section describes the Flash-CNNCap pipeline from layout rasterization to conductor-level capacitance prediction (Figure~\ref{fig:end_to_end}).  We first define the input encoding, then the dense prediction architecture, and finally the aggregation and supervision scheme used for training and inference.

\subsection{Input Encoding}
\label{sec:input_encoding}

Flash-CNNCap operates on the same fixed $224{\times}224$ per-layer raster representation introduced by 3D CNN-Cap~\cite{yang2022cnncap} (stages 1--2 in Figure~\ref{fig:end_to_end}).  Each extraction window is a square region of side length $S$ whose geometry is rasterized onto this grid, giving a per-pixel resolution of $S/224$.  Table~\ref{tab:dataset_params} lists the window sizes, resulting pixel resolutions, and minimum wire widths for different PDKs and window sizes.

Within each window, geometry is rasterized via a three-pass GPU pipeline into an $L{\times}224{\times}224$ local-ID map $M_w$, where $L$ is the number of metal and via layers.  First, a CUDA kernel paints each conductor rectangle into the appropriate layer of $M_w$ with its local conductor ID.  Second, a per-pixel metadata kernel scans $M_w$ to produce an occupancy mask and atomically count occupied pixels per conductor.  Third, a scatter kernel records the flat index of each occupied pixel into a per-conductor sparse index table $S_{w,\ell}$, which stores the positions in the flattened $L \cdot 224 \cdot 224$ volume that belong to conductor $\ell$.  These sparse indices are later consumed by the reduction kernel (Section~\ref{sec:aggregation}) without further host--device synchronization.

The resulting occupancy maps are encoded as input features.  For the total-capacitance model, each pixel is simply 1 if a conductor is present and 0 otherwise.  For the coupling model, the selected master conductor is encoded as $-1$ rather than $+1$, creating a three-valued encoding.  Formally, for each layer $\lambda$ and pixel $p$:
\begin{equation}
x^{\mathrm{tot}}_{w,\lambda}(p)=\mathbf{1}[m_{w,\lambda}(p) > 0],
\end{equation}
\begin{equation}
x^{\mathrm{env}}_{w,m,\lambda}(p)=\begin{cases}
-1 & \text{if } m_{w,\lambda}(p) = m,\\[2pt]
+1 & \text{if } m_{w,\lambda}(p) > 0 \text{ and } m_{w,\lambda}(p) \neq m,\\[2pt]
\phantom{+}0  & \text{otherwise},
\end{cases}
\end{equation}
where $m_{w,\lambda}(p)$ is the conductor ID at pixel $p$.  This encoding tells the coupling model which conductor acts as the source without removing the surrounding geometry that determines the target distribution.

Binary occupancy discards sub-pixel coverage; fractional-density inputs remain compatible and may reduce rasterization loss at coarser physical resolution.

\begin{table}[h!]
\centering
\caption{Window size, pixel resolution, and minimum wire width for each evaluated dataset.}
\label{tab:dataset_params}
\begin{tabular}{@{}llrrr@{}}
\toprule
\multirow{2}{*}{\textbf{Dataset}} & \multirow{2}{*}{\textbf{PDK}} & \textbf{Window} & \textbf{Min.\ width} & \textbf{Pixel} \\
 & & ($\mu$m) & (nm) & (nm) \\
\midrule
CNN-Cap & --- & 5.0 & 54 & 22.3 \\
\midrule
\multirow{2}{*}{Small}
    & NanGate45 & 2.0  &  65 &  8.9 \\
    & Sky130HD  &  4.5 & 140 & 20.1 \\
\midrule
\multirow{2}{*}{Medium}
    & NanGate45 & 5.0  &  65 & 22.3 \\
    & Sky130HD  & 10.0 & 140 & 44.6 \\
\midrule
\multirow{2}{*}{Large}
    & NanGate45 & 10.0 &  65 & 44.6 \\
    & Sky130HD  & 20.0 & 140 & 89.3 \\
\bottomrule
\end{tabular}
\end{table}

\subsection{Model Architecture}

The architectural change relative to CNN-Cap is in the prediction target (stage 3 in Figure~\ref{fig:end_to_end}).  A total-capacitance model predicts a dense, non-negative contribution tensor of shape $L{\times}224{\times}224$, producing one contribution channel per layer.  Summing this tensor over all layers and all pixels belonging to each conductor yields total capacitance for every conductor in one pass.  A second, master-conditioned coupling model predicts a contribution tensor of the same shape for all remaining conductors, recovering all pairwise couplings from that master in a single evaluation.  Full-matrix reconstruction therefore requires $1 + n$ forward passes for a window with $n$ conductors, compared with the $n + n(n{-}1)/2 = O(n^{2})$ passes required by the original image-to-scalar formulation.  A dense intermediate representation also preserves the spatial support of each conductor until the final reduction, so one forward pass can serve many conductor-level queries within the same window instead of collapsing the geometry into a single scalar too early.

\begin{figure}[!t]
  \centering
  \includegraphics[width=\linewidth]{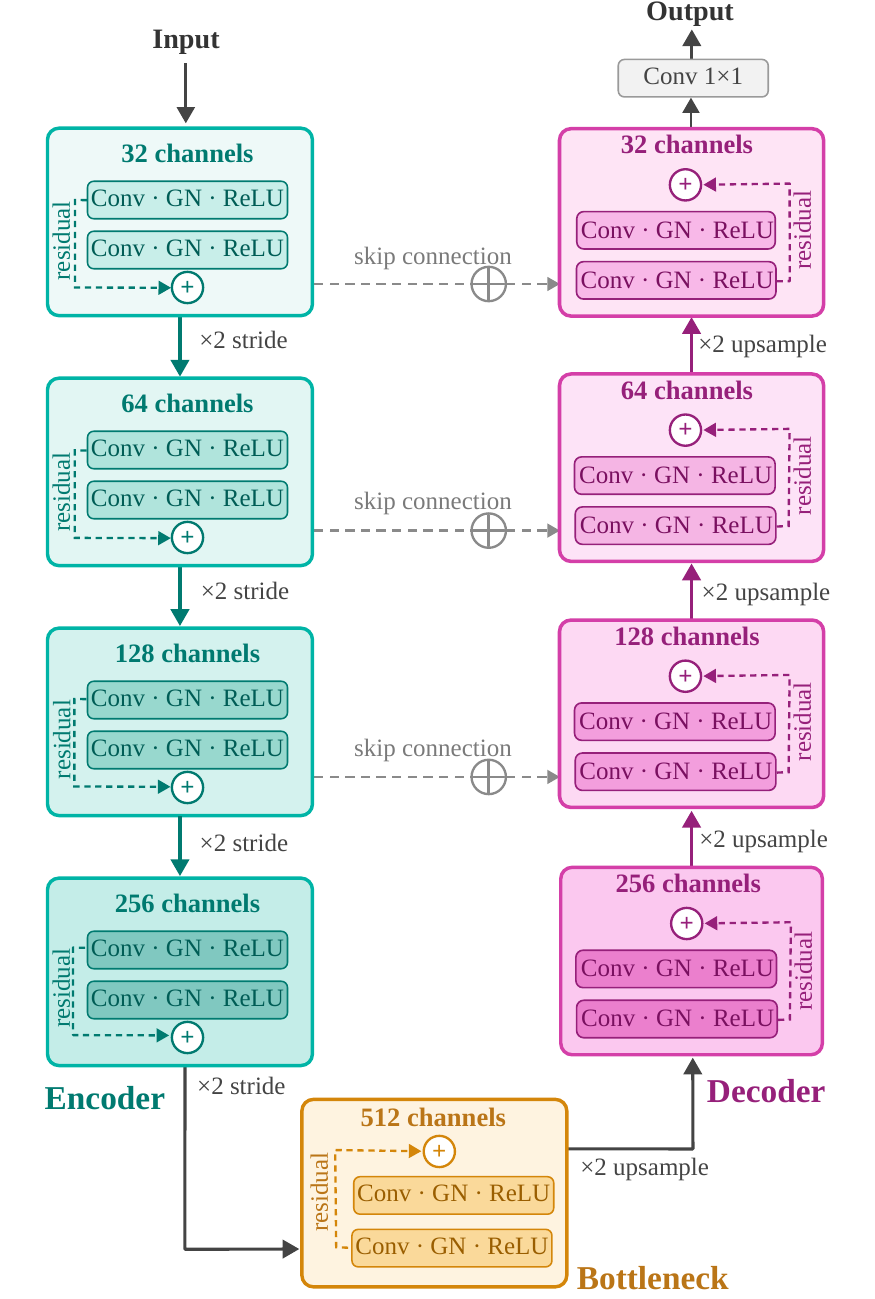}
  \caption{Selected D4\_D\_k5 U-Net architecture with four encoder/decoder levels, skip connections, and a 512-channel bottleneck.}
  \Description{Encoder--decoder block diagram with four downsampling and four upsampling stages connected by skip connections, converging at a 512-channel bottleneck.}
  \label{fig:unet_architecture}
\end{figure}

Because the output remains spatial, an encoder--decoder backbone is needed rather than a classification-style network.  Flash-CNNCap uses MONAI's residual 2D U-Net implementation~\cite{cardoso2022monai}, which builds on the original U-Net architecture~\cite{ronneberger2015unet}.  The model uses GroupNorm~\cite{wu2018groupnorm}, residual connections within each encoder and decoder block, bilinear upsampling followed by convolution, and $5{\times}5$ kernels.  The final $1{\times}1$ convolution produces $L$ channels of raw logits; an element-wise softplus ($\log(1 + e^x)$) maps them to strictly positive contribution values.  The non-negativity constraint is physically motivated: capacitance contributions are non-negative quantities, and enforcing this in the output space constrains the learned representation to match the additive structure of the underlying physics.  Skip connections allow fine routing geometry and layer-local context to bypass the bottleneck and remain available during decoding.  This preserves information that would be lost by the global average pooling used in ResNet-based scalar heads.  The architecture uses four encoder levels with channel widths $(32, 64, 128, 256, 512)$ and two residual units per block; Section~\ref{sec:ablation} evaluates alternatives.  As in CNN-Cap, total and coupling capacitance are handled by separate model instances trained independently; only the prediction granularity changes from scalar to spatial.  Figure~\ref{fig:unet_architecture} shows the selected architecture.

\subsection{Capacitance Aggregation and Supervision}
\label{sec:aggregation}

Once inference produces a contribution tensor, sparse reduction (stage 4 in Figure~\ref{fig:end_to_end}) aggregates it into conductor-level capacitances.  The total-capacitance model predicts a contribution tensor $q^{\mathrm{tot}}_w \in \mathbb{R}^{L \times H \times W}$ over the full window, where each element is the softplus-activated output.  The coupling model, conditioned on a master conductor $m$, predicts $q^{\mathrm{env}}_{w,m} \in \mathbb{R}^{L \times H \times W}$ over all environmental conductors.  Let $m_w(p)$ denote the local conductor ID at pixel $p$ (with $0$ for background), $\pi_w(\ell)$ the mapping to global net index, and $I_{w,\ell} = \{p \mid m_w(p) = \ell\}$ the pixel set for conductor $\ell$.  No ground-truth pixel maps are required.  Algorithm~\ref{alg:aggregation} summarizes the inference-time aggregation procedure.
\begin{algorithm}[!b]
\caption{GPU-resident capacitance extraction pipeline.}
\label{alg:aggregation}
\begin{algorithmic}[1]
\State Initialize $\hat{C}^{\mathrm{tot}}_n \leftarrow 0$,\;
       $\hat{C}^{\mathrm{dir}}_{i \rightarrow j} \leftarrow 0$
       for all nets $n$, $i$, $j$
\For{each window $w \in \mathcal{W}$}
  \State \textbf{Rasterize} (GPU): paint rectangles $\rightarrow$ local-ID map $M_w \in \mathbb{Z}^{L \times H \times W}$;
         build occupancy mask and per-conductor sparse index tables $S_{w,\ell}$
  \State \textbf{Total model}: $q^{\mathrm{tot}}_w \leftarrow \mathrm{softplus}(f_{\theta}(\mathbf{1}[M_w > 0]))$
  \State \textbf{Sparse reduce} (GPU): for each conductor $\ell$,\;
         $\hat{C}^{\mathrm{tot}}_{\pi_w(\ell)}
         \mathrel{+}= \sum_{k} q^{\mathrm{tot}}_w[S_{w,\ell}(k)]$
  \For{each visible master conductor $m \in V_w$}
    \State Build env features: $x_m \leftarrow \mathbf{1}[M_w > 0] - 2\,\mathbf{1}[M_w = m]$
    \State $q^{\mathrm{env}}_{w,m} \leftarrow \mathrm{softplus}(g_{\phi}(x_m))$
    \State \textbf{Sparse reduce} (GPU): for each target $\ell \neq m$,\;
           $\hat{C}^{\mathrm{dir}}_{\pi_w(m) \rightarrow \pi_w(\ell)}
           \mathrel{+}= \sum_{k} q^{\mathrm{env}}_{w,m}[S_{w,\ell}(k)]$
  \EndFor
\EndFor
\State \textbf{Symmetrize:} $\hat{C}^{\mathrm{cpl}}_{\{i,j\}} =
       \tfrac{1}{2}(\hat{C}^{\mathrm{dir}}_{i \rightarrow j} +
       \hat{C}^{\mathrm{dir}}_{j \rightarrow i})$ for all $i \neq j$
\end{algorithmic}
\end{algorithm}

On the GPU, the reduction uses the per-conductor sparse index tables $S_{w,\ell}$ built during rasterization.  The contribution tensor is flattened to a vector of length $L \cdot H \cdot W$, and a dedicated CUDA kernel sums the entries at the sparse indices for each conductor in a single pass:
\begin{equation}
\hat{C}_{w,\ell} = \sum_{k=1}^{|S_{w,\ell}|} q_w[S_{w,\ell}(k)].
\end{equation}
Each conductor's reduction is independent and assigned to a separate GPU thread, yielding $O(1)$ kernel launches per window regardless of conductor count.  This avoids materializing the full dense map on the host and eliminates the $O(L \cdot H \cdot W)$ per-conductor scan that a naive dense reduction would require. Written at the conductor level, the readout used by both training and inference is
\begin{equation}
\hat{C}^{\mathrm{tot}}_{w,\ell} = \sum_{p \in I_{w,\ell}} q^{\mathrm{tot}}_w(p), \qquad
\hat{C}^{\mathrm{dir}}_{w,m \rightarrow \ell} = \sum_{p \in I_{w,\ell}} q^{\mathrm{env}}_{w,m}(p).
\end{equation}
with a final symmetrization
\begin{equation}
\hat{C}^{\mathrm{cpl}}_{\{i,j\}} = \tfrac{1}{2}\left(\hat{C}^{\mathrm{dir}}_{i \rightarrow j} + \hat{C}^{\mathrm{dir}}_{j \rightarrow i}\right).
\end{equation}
Under the standard Maxwell-matrix sign convention, the reconstructed matrix entries are
\begin{equation}
\hat{M}_{ij} =
\begin{cases}
\hat{C}^{\mathrm{tot}}_i & \text{if } i = j,\\
-\hat{C}^{\mathrm{cpl}}_{\{i,j\}} & \text{if } i \neq j.
\end{cases}
\end{equation}
Softplus makes predicted totals and couplings non-negative; symmetrization preserves this, while the Maxwell convention supplies negative off-diagonal signs.
For a set of query conductors $Q$, training minimizes masked MSRE,
\begin{equation}
\mathcal{L}_{\mathrm{MSRE}} = \frac{1}{|Q|} \sum_{j \in Q}
\left(\frac{\hat{C}_j - C_j}{C_j + \epsilon}\right)^2,
\end{equation}
which matches the evaluation emphasis on relative rather than absolute error.  This objective is especially important for coupling capacitance, where values span a wide dynamic range and small couplings would otherwise be underweighted by an absolute-error loss.

\begin{figure*}[!b]
  \centering
  \includegraphics[width=0.9\textwidth]{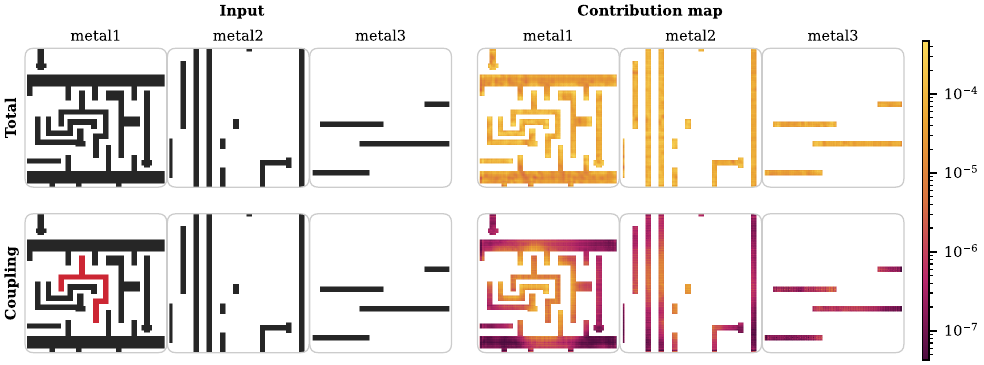}
  \caption{Learned contribution maps on a NanGate45 window. Top: total capacitance; bottom: coupling capacitance, with the master conductor shown in red. Brighter values indicate larger per-pixel contributions.}
  \Description{Six-panel visualization showing input occupancy maps and corresponding predicted contribution maps for total and coupling capacitance across three metal layers of a NanGate45 window.}
  \label{fig:contribution_maps}
\end{figure*}

Conductor-level supervision suffices because capacitances are conductor-wise and the rasterized ID map defines exact conductor support.  The model thus learns a non-unique, non-negative latent decomposition whose masked sums match labeled parasitics.  Figure~\ref{fig:contribution_maps} illustrates a representative NanGate45 window.  Background pixels receive no gradient because they are excluded from conductor masks; the structured patterns are therefore latent representations, not exact physical field solutions.

\section{Results}

This section evaluates Flash-CNNCap under a consistent training and benchmarking protocol.  We first summarize the experimental setup and then present the ablation, runtime, accuracy, and end-to-end pipeline results.

\subsection{Experimental Setup}
\label{sec:setup}

Experiments were run on a server with 2 Intel Xeon Platinum 8488C CPUs (96 cores), 251\,GiB RAM, and 8 NVIDIA GeForce RTX 5090 GPUs. All neural inference measurements use one RTX 5090 GPU.  OpenRCX is run in its default single-threaded configuration, and RWCap uses 8 CPU cores for the end-to-end timing comparison.

All models are trained for 100 epochs with AdamW~\cite{loshchilov2019adamw}, learning rate $3{\times}10^{-4}$, weight decay $10^{-4}$, batch size 16, and MSRE loss.  We set $(\beta_1,\beta_2)=(0.9,0.999)$, use a 5-epoch linear warmup, decay the learning rate to 10\% of the base value with a cosine schedule, and clip gradients at max norm 1.0.  The fixed-seed 80/20 split is window-level, not target-level.  It is window-disjoint but not design-disjoint: no instance occurs in both sets, although distinct windows from one routed design may.  Thus, the results measure window-level generalization within CapBench.

All models are evaluated using mean absolute relative error (MARE) as the primary accuracy metric.  We additionally report the fraction of predictions exceeding 5\% and 10\% relative error to characterize the error distribution beyond the mean.

\begin{table}[t]
\centering
\caption{CNN-Cap reproduction (MARE).}
\label{tab:cnncap_legacy_compare}
\begin{tabular}{lrrr}
\toprule
\multirow{2}{*}{\textbf{Target}} & \multicolumn{3}{c}{\textbf{ResNet-34}} \\
\cmidrule(lr){2-4}
& Checkpoint & Reproduction & Proposed \\
\midrule
Total    & 0.0112 & 0.0165 & \cellcolor[gray]{0.9}0.0067 \\
Coupling & 0.0316 & 0.0368 & \cellcolor[gray]{0.9}0.0121 \\
\bottomrule
\end{tabular}
\end{table}

Table~\ref{tab:cnncap_legacy_compare} is a protocol check before the dense-model ablation.  With the scalar CNN-Cap architecture held fixed at ResNet-34, the proposed training recipe improves over both the released checkpoint and a best-effort reproduction from the paper description.

\subsection{Ablation Study}
\label{sec:ablation}

\begin{figure}[b]
  \centering
  \includegraphics[width=\linewidth]{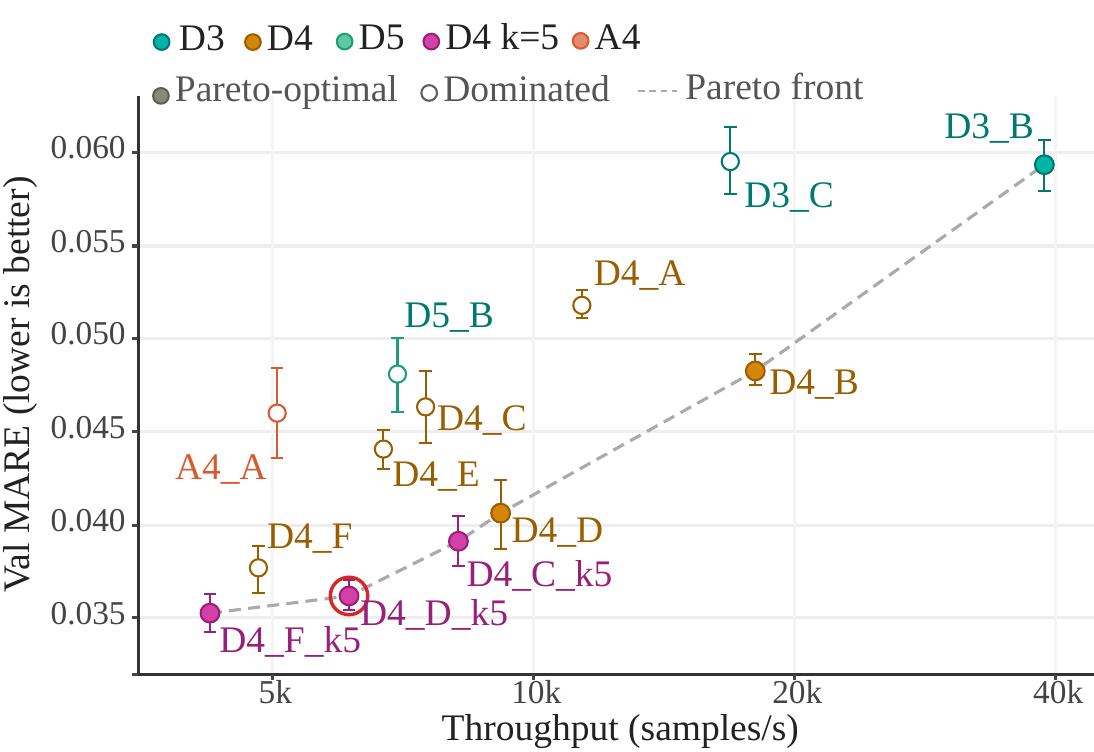}
  \caption{Accuracy vs.\ throughput Pareto analysis on NanGate45 small. Error bars show 95\% CI across five seeds.}
  \Description{Scatter plot of validation MARE versus TensorRT throughput for 13 U-Net variants with Pareto-optimal points highlighted, 95 percent confidence intervals, and the selected model surrounded by a red circle.}
  \label{fig:pareto}
\end{figure}

\begin{table*}[!t]
\centering
\caption{Ablation on NanGate45 small (5 seeds, 100 epochs). D-series: MONAI 2D U-Net; A4\_A: Attention U-Net. Throughput measured with TensorRT at batch 64. Shaded row: selected model.}

\label{tab:ablation}
\begin{tabular}{lcccl rrrr}
\toprule
\multirow{2}{*}{\textbf{Model}} &
\multirow{2}{*}{\textbf{Depth}} &
\multirow{2}{*}{\textbf{Kernel}} &
\multirow{2}{*}{\textbf{Res. units}} &
\multirow{2}{*}{\textbf{Channels}} &
\textbf{Params} &
\textbf{GFLOPs} &
\textbf{Samp/s} &
\textbf{Val MARE} \\
& & & & & (M) & /sample & (k) & $\mu \pm \sigma$ \\
\midrule
D3\_B     & 3 & 3 & 0          & (16,32,64,128)         & 0.16  &  0.37 & 38.9 & $0.0593 \pm 0.0011$ \\
D3\_C     & 3 & 3 & 1          & (16,32,64,128)         & 0.21  &  0.59 & 16.9 & $0.0595 \pm 0.0014$ \\
\midrule
D4\_A     & 4 & 3 & 0          & (24,48,96,192,384)     & 1.49  &  0.98 & 11.4 & $0.0518 \pm 0.0006$ \\
D4\_B     & 4 & 3 & 0          & (32,64,128,256,512)    & 2.64  &  1.72 & 18.0 & $0.0483 \pm 0.0007$ \\
D4\_C     & 4 & 3 & 1          & (32,64,128,256,512)    & 3.36  &  2.86 &  7.5 & $0.0463 \pm 0.0016$ \\
D4\_D     & 4 & 3 & 2          & (32,64,128,256,512)    & 6.50  &  4.72 &  9.2 & $0.0406 \pm 0.0015$ \\
D4\_E     & 4 & 3 & 1          & (48,96,192,384,768)    & 7.54  &  6.31 &  6.7 & $0.0441 \pm 0.0008$ \\
D4\_F     & 4 & 3 & 2          & (48,96,192,384,768)    & 14.62 & 10.49 &  4.8 & $0.0377 \pm 0.0010$ \\
D4\_C\_k5 & 4 & 5 & 1          & (32,64,128,256,512)    & 9.08  &  7.82 &  8.2 & $0.0391 \pm 0.0011$ \\
\rowcolor[gray]{0.9}
D4\_D\_k5 & 4 & 5 & 2          & (32,64,128,256,512)    & 17.81 & 12.96 &  6.1 & $0.0362 \pm 0.0006$ \\
D4\_F\_k5 & 4 & 5 & 2          & (48,96,192,384,768)    & 40.07 & 28.86 &  4.2 & $0.0353 \pm 0.0008$ \\
\midrule
D5\_B     & 5 & 3 & 1          & (24,48,96,192,384,768) & 7.59  &  2.03 &  7.0 & $0.0481 \pm 0.0016$ \\
\midrule
A4\_A     & 4 & 3 & \textemdash & (32,64,128,256,512)   & 7.94  & 16.41 &  5.1 & $0.0460 \pm 0.0019$ \\
\bottomrule
\end{tabular}
\end{table*}

The ablation evaluates 13 dense-prediction backbones on the NanGate45 small split under a fixed training protocol (100 epochs, AdamW, MSRE loss, five random seeds).  We sweep MONAI U-Net depth (D3--D5), channel width, residual depth, and kernel size ($3$ vs.\ $5$), and we include an Attention U-Net comparator.  The Attention U-Net does not improve over standard U-Nets, indicating that skip connections already provide sufficient long-range context for parasitic layouts.  Moving from three to four encoder levels consistently reduces MARE, while five-level models show no further gain.  Within the four-level family, additional residual depth and larger $5{\times}5$ kernels are both beneficial.

Figure~\ref{fig:pareto} and Table~\ref{tab:ablation} show that the useful Pareto frontier is narrow and is dominated by four-level U-Nets.  D4\_F\_k5 attains the lowest MARE in the sweep, but the absolute gain over D4\_D\_k5 is only 0.0009 while FLOPs more than double and TensorRT throughput drops from 6.1k to 4.2k samples/s.  D4\_D\_k5 therefore captures most of the accuracy benefit of the larger models without paying their full deployment cost, which is why it is used in the remaining experiments.

\begin{table}[b]
\centering
\caption{Inference passes for full-matrix extraction (256 windows).}
\label{tab:windows}
\begin{tabular}{l rr r}
\toprule
\textbf{Dataset} & \textbf{Image-to-scalar} & \textbf{Image-to-image} & \textbf{Reduction} \\
\midrule
Small  &    39,346 &  4,316 &  9.1$\times$ \\
Medium &   400,556 & 13,737 & 29.2$\times$ \\
Large  & 2,401,707 & 34,540 & 69.5$\times$ \\
\bottomrule
\end{tabular}
\end{table}

\subsection{Inference Time}

\begin{figure}[b]
  \centering
  \includegraphics[width=\linewidth]{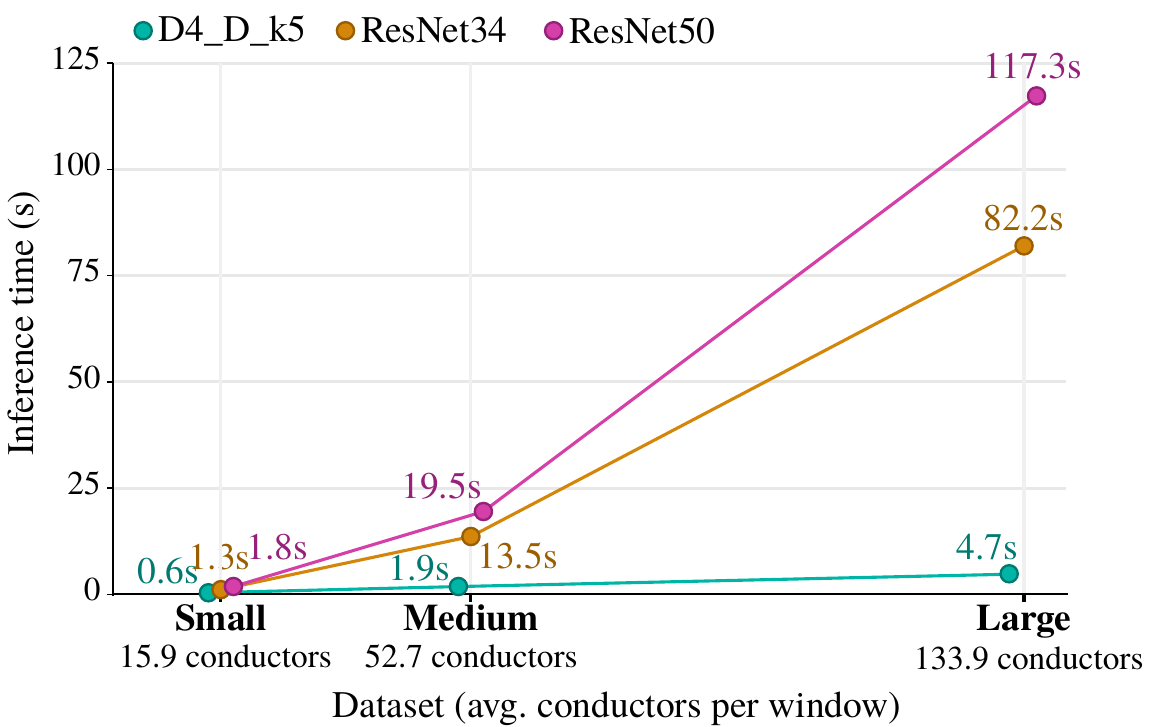}
  \caption{Full-matrix inference time over 256 windows on one RTX 5090 GPU.}
  \Description{Bar chart comparing full-matrix inference time across small, medium, and large datasets for D4\_D\_k5, ResNet-34, and ResNet-50.}
  \label{fig:inference_time}
\end{figure}

Figure~\ref{fig:inference_time} reports the full-matrix reconstruction time over 256 windows on one RTX 5090 GPU, including GPU forward passes and, for Flash-CNNCap, the sparse reduction over spatial contribution maps.  On the small set (15.9 conductors/window on average), Flash-CNNCap takes 0.6\,s versus 1.3--1.8\,s for the ResNet baselines.  On the medium set (52.7 conductors/window), the gap widens to 1.9\,s versus 13.5--19.5\,s.  On the large set (133.9 conductors/window), Flash-CNNCap finishes in 4.7\,s while ResNet-34 and ResNet-50 require 82.2\,s and 117.3\,s, yielding a measured $17.5\times$ same-GPU speedup enabled primarily by the reduction in model evaluations.  Each U-Net pass is heavier than a scalar-regression pass, but the pass-count reduction dominates at scale.

Table~\ref{tab:windows} explains the crossover.  On the same 256-window evaluation, Flash-CNNCap reduces the number of model calls by $9.1\times$ on small windows, $29.2\times$ on medium windows, and $69.5\times$ on large windows.  Once conductor counts grow, the pass-count reduction dominates despite the heavier per-pass cost.

\begin{table*}[t]
\centering
\caption{Validation MARE and fraction of predictions exceeding 5\% and 10\% relative error. Bold: best MARE per row.}
\label{tab:results_full}
\begin{tabular}{lll rrr rrr rrr}
\toprule
& & &
\multicolumn{3}{c}{\textbf{ResNet-34}} &
\multicolumn{3}{c}{\textbf{ResNet-50}} &
\multicolumn{3}{c}{\textbf{D4\_D\_k5}} \\
\cmidrule(lr){4-6} \cmidrule(lr){7-9} \cmidrule(lr){10-12}
\textbf{Dataset} & \textbf{PDK} & \textbf{Target} &
MARE & $>$5\% & $>$10\% &
MARE & $>$5\% & $>$10\% &
MARE & $>$5\% & $>$10\% \\
\midrule
\multirow[c]{2}{*}{CNN-Cap} & \multirow[c]{2}{*}{---}
  & Total    & \cellcolor[gray]{0.9}0.0067 & 0.0022 & 0.0007 & \cellcolor[gray]{0.9}0.0067 & 0.0022 & 0.0000 & 0.0072 & 0.0096 & 0.0000 \\
  &
  & Coupling & \cellcolor[gray]{0.9}0.0121 & 0.1121 & 0.0313 & 0.0248 & 0.1199 & 0.0353 & 0.0278 & 0.1401 & 0.0423 \\
\midrule
\multirow[c]{4}{*}{Small} & \multirow[c]{2}{*}{NanGate45}
  & Total    & 0.0252 & 0.0747 & 0.0114 & \cellcolor[gray]{0.9}0.0223 & 0.0600 & 0.0100 & 0.0287 & 0.1099 & 0.0301 \\
  &
  & Coupling & 0.0566 & 0.3908 & 0.1403 & 0.0575 & 0.3784 & 0.1437 & \cellcolor[gray]{0.9}0.0330 & 0.1950 & 0.0393 \\
\cmidrule{2-12}
& \multirow[c]{2}{*}{Sky130HD}
  & Total    & 0.0309 & 0.1233 & 0.0459 & 0.0419 & 0.2137 & 0.0768 & \cellcolor[gray]{0.9}0.0205 & 0.0766 & 0.0184 \\
  &
  & Coupling & 0.0564 & 0.3519 & 0.1354 & 0.0619 & 0.3748 & 0.1593 & \cellcolor[gray]{0.9}0.0363 & 0.2194 & 0.0566 \\
\midrule
\multirow[c]{4}{*}{Medium} & \multirow[c]{2}{*}{NanGate45}
  & Total    & 0.0207 & 0.0610 & 0.0162 & 0.0172 & 0.0477 & 0.0109 & \cellcolor[gray]{0.9}0.0171 & 0.0518 & 0.0096 \\
  &
  & Coupling & 0.0442 & 0.2803 & 0.0861 & 0.0424 & 0.2714 & 0.0778 & \cellcolor[gray]{0.9}0.0333 & 0.2138 & 0.0405 \\
\cmidrule{2-12}
& \multirow[c]{2}{*}{Sky130HD}
  & Total    & 0.0151 & 0.0376 & 0.0054 & \cellcolor[gray]{0.9}0.0142 & 0.0363 & 0.0054 & 0.0146 & 0.0405 & 0.0066 \\
  &
  & Coupling & 0.0357 & 0.2342 & 0.0520 & 0.0358 & 0.2345 & 0.0518 & \cellcolor[gray]{0.9}0.0302 & 0.1840 & 0.0323 \\
\midrule
\multirow[c]{4}{*}{Large} & \multirow[c]{2}{*}{NanGate45}
  & Total    & 0.0277 & 0.1203 & 0.0341 & \cellcolor[gray]{0.9}0.0246 & 0.0948 & 0.0234 & 0.0294 & 0.1546 & 0.0420 \\
  &
  & Coupling & 0.0593 & 0.3801 & 0.1510 & 0.0516 & 0.3394 & 0.1150 & \cellcolor[gray]{0.9}0.0456 & 0.3143 & 0.0900 \\
\cmidrule{2-12}
& \multirow[c]{2}{*}{Sky130HD}
  & Total    & 0.0323 & 0.1439 & 0.0430 & \cellcolor[gray]{0.9}0.0230 & 0.1043 & 0.0232 & 0.0310 & 0.1712 & 0.0506 \\
  &
  & Coupling & 0.0557 & 0.3852 & 0.1481 & 0.0521 & 0.3638 & 0.1307 & \cellcolor[gray]{0.9}0.0447 & 0.3143 & 0.0971 \\
\bottomrule
\end{tabular}
\end{table*}

\begin{figure*}[b]
  \centering
  \includegraphics[width=0.9\linewidth]{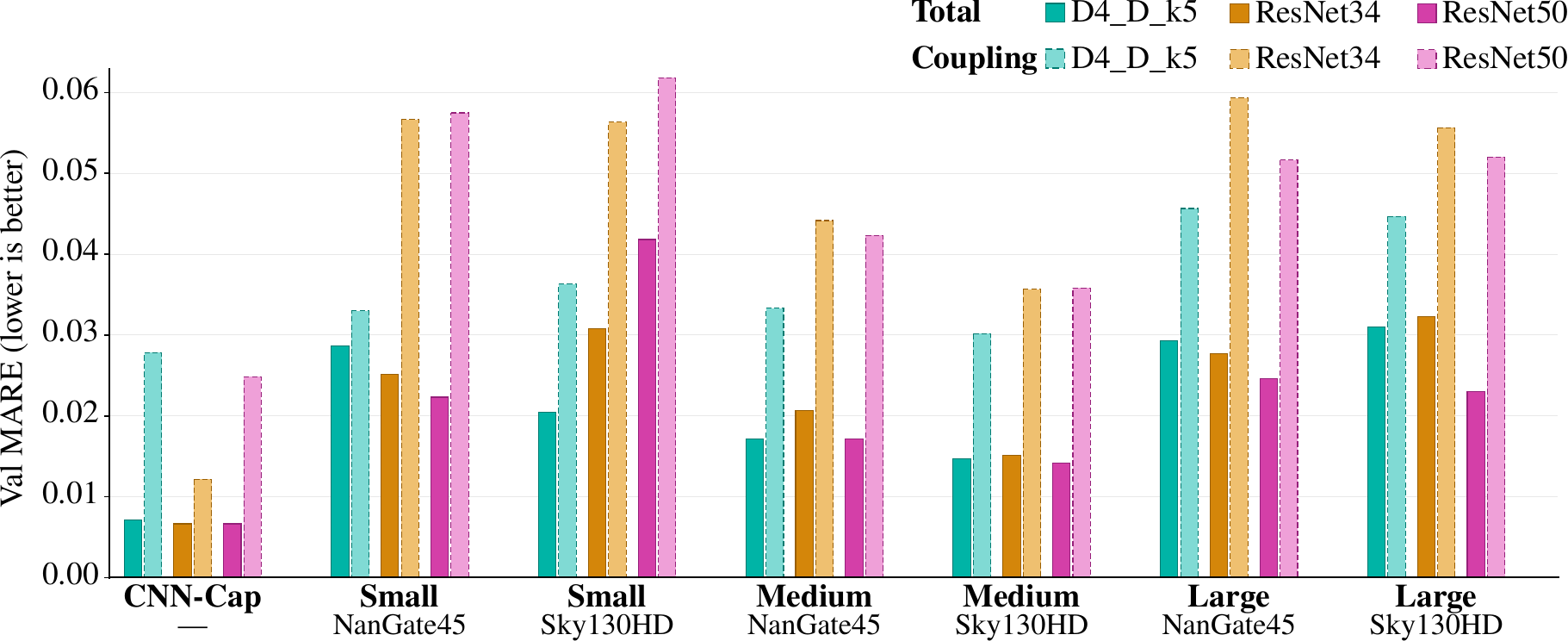}
  \caption{Validation MARE across all benchmarks (100 epochs).}
  \Description{Grouped bar chart showing total and coupling MARE for three models across the CNN-Cap legacy set and six CapBench subsets.}
  \label{fig:accuracy_comparison}
\end{figure*}

\subsection{Accuracy Comparisons}

As a setup sanity check on the legacy CNN-Cap task (Table~\ref{tab:cnncap_legacy_compare}), the released ResNet-34 checkpoint yields 1.12\%/3.16\% MARE for total/coupling on our validation split, consistent with the published figures of 1.10\%/3.10\%.  A best-effort reproduction based on the paper description reaches 1.65\%/3.68\%, while the updated training recipe from Section~\ref{sec:setup} improves these to 0.67\%/1.21\%.  Since the original training code was not released, we reconstructed the pipeline from the paper; the improvement should therefore be attributed to the training and input recipe rather than to architectural changes.  Notably, both the released checkpoint and our reproduction use CNN-Cap's original per-layer density maps as input, whereas the updated recipe uses binary occupancy masks (Section~\ref{sec:input_encoding}).  Matching or surpassing the density-map baselines establishes binary occupancy as the submitted operating point, rather than showing that fractional-density inputs are generally unnecessary.

On the CapBench subsets (Table~\ref{tab:results_full} and Figure~\ref{fig:accuracy_comparison}), ResNet-50 and D4\_D\_k5 are competitive on total capacitance across subsets, with all three models remaining in the same accuracy regime.  The threshold metrics reinforce this split: the best total-capacitance model depends on the dataset; for coupling capacitance, however, D4\_D\_k5 achieves the lowest MARE and the lowest fractions above both 5\% and 10\% relative error on every evaluated CapBench subset.

This accuracy trade-off does not require higher training cost: D4\_D\_k5 training times were consistently lower than or comparable to the ResNet baselines (e.g., 16.7--45.7\,h vs.\ 16.6--51.6\,h for coupling on small subsets).

The contrast between total and coupling accuracy reflects the structure of the prediction task.  Because total capacitance integrates over the full conductor support, a scalar backbone captures it effectively once it encodes occupancy, layer context, and conductor extent.  Coupling capacitance is more demanding: for a marked master conductor, the model must preserve how the contribution is partitioned across many visible targets.  Dense prediction matches that target more naturally because the conductor structure is kept explicit until the final masked reduction.  The gain is a systematic tightening of the coupling-error distribution.

Since full-matrix reconstruction is dominated by pairwise coupling terms, this distinction matters more than a small difference in the total-capacitance results.  The dense model's main advantage is that it is simultaneously the fastest full-matrix extractor and the most reliable coupling predictor across all evaluated CapBench subsets.

\subsection{Full Pipeline}
\label{sec:pipeline}

An end-to-end extractor must prepare routed geometry, execute and reduce model predictions, and emit standard parasitic output.

Flash-CNNCap uses the same standards-compatible DEF/SPEF interface as OpenRCX~\cite{OpenRCX2021}.  Python orchestrates C++ geometry parsing and SPEF writing on the CPU; custom CUDA extensions keep rasterization, occupancy construction, TensorRT inference (batch size 24), and sparse reduction device-resident.  Each symmetrized term $\hat{C}^{\mathrm{cpl}}_{\{i,j\}}$ is written as a coupling entry, with $\hat{C}^{\mathrm{gnd}}_i = \hat{C}^{\mathrm{tot}}_i - \sum_{j \neq i} \hat{C}^{\mathrm{cpl}}_{\{i,j\}}$.  Independent total and coupling training does not guarantee a non-negative derived grounded term.  Contribution maps remain on the GPU, where the input ID map is reused for conductor-level reduction.

\begin{figure}[b]
  \centering
  \includegraphics[width=\linewidth]{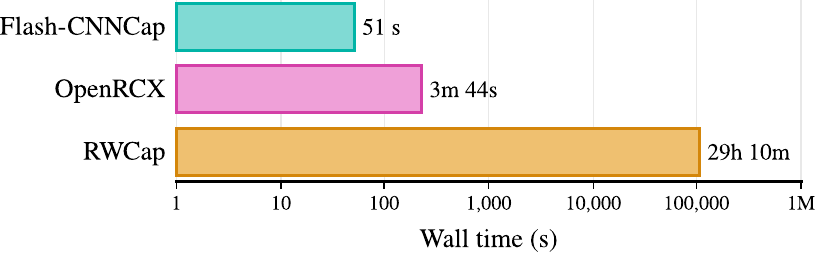}
  \caption{End-to-end runtime on 1{,}024 NanGate45 large windows (Flash-CNNCap on 1 RTX 5090 GPU, OpenRCX single-threaded, RWCap on 8 CPU cores).}
  \Description{Horizontal bar chart comparing end-to-end wall time for RWCap, OpenRCX, and Flash-CNNCap on 1024 NanGate45 large windows.}
  \label{fig:timing_comparison}
\end{figure}

Dense outputs increase activation memory; the deployed TensorRT batch size of 24 can be reduced without changing the $O(n)$ model-evaluation count.  Compact GPU-resident buffers and no inner-loop host--device synchronization limit overhead.  Fixed windows may truncate long-range boundary interactions; overlapping windows with an interior retention margin are a possible, unevaluated mitigation.

Table~\ref{tab:pipeline} breaks down the measured runtime on 1{,}024 NanGate45 large windows.  The combined GPU path accounts for 70.4\% of the total 51.23\,s; SPEF writing contributes 26.8\%.  With full-matrix reconstruction reduced to a linear number of model evaluations, end-to-end throughput is governed by the efficiency of the device-side path and output generation, not by neural inference alone.

\begin{table}[t]
\centering
\caption{Runtime breakdown on 1{,}024 NanGate45 large windows.}
\label{tab:pipeline}
\begin{tabular}{llrr}
\toprule
\textbf{Stage} & \textbf{Backend} & \textbf{Time (s)} & \textbf{Share (\%)} \\
\midrule
Initialization  & Python/C++/TRT & 1.31 & 2.6 \\
DEF parsing     & CPU (C++)      & 0.12 & 0.2 \\
GPU path        & CUDA/TRT       & 36.06 & 70.4 \\
SPEF writing    & CPU (C++)      & 13.75 & 26.8 \\
\midrule
\textbf{Total}  &                & \textbf{51.23} & \textbf{100.0} \\
\bottomrule
\end{tabular}
\end{table}

Figure~\ref{fig:timing_comparison} compares one-GPU Flash-CNNCap against RWCap~\cite{yu2013rwcap} on 8 CPU cores and single-threaded OpenRCX for the same 1{,}024 windows: 51.23\,s, 104{,}982.99\,s, and 223.79\,s, respectively.  This heterogeneous systems reference is not hardware-normalized; the same-GPU scalar-CNN comparison appears in the inference-time evaluation.  RWCap primarily supplies accuracy labels, whereas OpenRCX is the practical comparator but solves a broader problem and emits richer RC information.

\section{Conclusions}

Flash-CNNCap reformulates CNN-based capacitance extraction as image-to-image regression over spatial contribution maps, reducing full-matrix reconstruction from quadratic to linear in the number of conductors while preserving the raster representation that makes CNNs effective on routed IC layouts.  The contribution maps are learned as a latent, conductor-mask-aggregated decomposition from conductor-level supervision rather than from per-pixel labels.  Experimentally, the approach matches scalar ResNet baselines on total capacitance, delivers the strongest coupling accuracy across all evaluated CapBench subsets, and provides an order-of-magnitude full-matrix speedup that grows with conductor density.  An end-to-end DEF-to-SPEF pipeline demonstrates that the speedup survives systems-level overheads and outperforms OpenRCX on the same benchmark.

Dense prediction is most valuable where the extractor must preserve a structured distribution over many outputs rather than collapse a window into a single scalar.  The same representation that improves target-wise decomposition also eliminates redundant pairwise inference, producing the largest accuracy gains on coupling capacitance while sustaining the speedup at the systems level.  Finite-window context and the lack of an explicit joint constraint on the derived grounded capacitance remain limitations.  The approach may apply to other post-layout learning problems in which pairwise or set-valued parasitics dominate runtime.


\clearpage
\bibliographystyle{ACM-Reference-Format-Cite-Order}
\bibliography{refs}

\end{document}